# An Interpretable Machine Learning Framework to Understand Bikeshare Demand before and during the COVID-19 Pandemic in New York City

**Majbah Uddin, PhD, Corresponding author**
National Transportation Research Center
Oak Ridge National Laboratory
1 Bethel Valley Road
Oak Ridge, TN 37830
ORCiD: 0000-0001-9925-3881
Email: uddinm@ornl.gov

**Ho-Ling Hwang, PhD**
National Transportation Research Center
Oak Ridge National Laboratory
1 Bethel Valley Road
Oak Ridge, TN 37830
ORCiD: 0000-0001-7651-6263
Email: hwanghlc@gmail.com

**Md Sami Hasnine, PhD**
Assistant Professor
Department of Civil and Environmental Engineering
Howard University
2300 Sixth Street, NW #1026
Washington, DC, USA, 20059
ORCiD: 0000-0003-4110-5047
Email: mdsami.hasnine@howard.edu

# An Interpretable Machine Learning Framework to Understand Bikeshare Demand before and during the COVID-19 Pandemic in New York City

In recent years, bikesharing systems have become increasingly popular as affordable and sustainable micromobility solutions. Advanced mathematical models such as machine learning are required to generate good forecasts for bikeshare demand. To this end, this study proposes a machine learning modeling framework to estimate hourly demand in a large-scale bikesharing system. Two Extreme Gradient Boosting models were developed: one using data from before the COVID-19 pandemic (March 2019 to February 2020) and the other using data from during the pandemic (March 2020 to February 2021). Furthermore, a model interpretation framework based on SHapley Additive exPlanations was implemented. Based on the relative importance of the explanatory variables considered in this study, share of female users and hour of day were the two most important explanatory variables in both models. However, the month variable had a higher importance in the pandemic model than in the pre-pandemic model.

**Introduction**

Bikesharing systems have become increasingly popular as affordable and sustainable micromobility solutions in recent years. In the United States, large cities such as New York City, San Francisco (SF), Washington DC, Chicago, and Boston experienced their highest bikeshare demands in 2020 and 2021. For example, according to the US Department of Transportation website accessed on July 5, 2022, at least 16.7 million trips were made at selected docked bikeshare systems in New York City from January to August 2021, 43% more than in January to August 2020. Although bikesharing ridership was already increasing exponentially before the COVID-19 pandemic began, the pandemic is largely responsible for the skyrocketing bikesharing demand. Considering only pandemic period, bikesharing demand decreased early in the pandemic (2020), but it increased later, particularly in 2021 (Wang and Noland 2021a).

Many previous studies have explored the factors that affect bikesharing demand, particularly sociodemographic factors, weather, and the built environment (Babagoli et al. 2019; El-Assi et al. 2017; Heaney et al. 2019; Reilly et al. 2020a; Wang and Noland 2021a,b; Xu and Chow 2020). A few recent studies have examined bikeshare trip characteristics during the pandemic (Basu and Ferreira 2021; Jobe and Griffin 2021; Padmanabhan et al. 2021). Another study examined equity concerns related to bikeshare demand (Ursaki and Aultman-Hall 2015). Because the majority of the described studies relied on either descriptive statistics or simple aggregate modeling techniques (e.g., linear regression), they are difficult for policymakers and planners to use for forecasting and policymaking. Understanding the factors that affect bikeshare demand is extremely important for urban planning and policymaking. Advanced mathematical models such as machine learning are required to generate good forecast for bikeshare demand, but many previous studies have avoided using machine learning models because of their limitations in terms of variable interpretability.

To fill this gap in bikeshare demand modeling, this study proposes a machine learning modeling framework to estimate hourly demand in a large-scale bikesharing system. The study used 39.9 million bikesharing transactions from March 2019 to February 2021 in New York City from Citi Bike system (CBS) data obtained from the Citi Bike website on July 31, 2022. Each transaction record includes details such as trip duration; bike check out and check in time; pick up and drop off station names, latitudes, and longitudes; and user ID. These records were aggregated by the hour to obtain the citywide demand during the selected time period. The aggregated data were then augmented with weather data and adjusted for holidays. The main contributions of this study are summarized here.

1. A machine learning modeling framework is proposed to estimate hourly demand in a large-scale bikesharing system. To isolate the effect of the pandemic on bikeshare demand, two Extreme Gradient Boosting (XGBoost) models were developed. One model uses data from before the pandemic (March 2019 to February 2020), and the other uses data from during the pandemic (March 2020 to February 2021).
2. To understand the "black box" of the machine learning model, a model interpretation framework based on SHapley Additive exPlanations (SHAP) was implemented. The model interpretations indicate how much each explanatory variable contributes, either positively or negatively, to bikeshare demand. Many variables were evaluated in this study that were not considered in most previous studies.

**Literature review**

Literature reviewed for this study was categorized by three themes: the effects of the pandemic, weather and built environment, and sociodemographic factors on bikeshare demand. These three themes are discussed briefly here. Table 1 summarizes the papers reviewed in this section.

*Effect of the COVID-19 pandemic on bikeshare demand*

A recent study found that pandemic-related regulations negatively affected the bikeshare demand in New York City (Wang and Noland 2021a). However, the pandemic was more detrimental to subway ridership than bikeshare ridership. In another study, Wang and Noland (2021b) showed that despite its initial drop, bikeshare demand in New York City reached its pre-pandemic level by the summer of 2020. Jobe and Griffin (2021) designed several surveys and collected user data in San Antonio, Texas to understand how bikeshare systems in the area responded to the pandemic. They found that only 5 out of 11 bikeshare companies actively worked to reduce COVID-19 transmission. Whereas Jobe and Griffin (2021) investigated stakeholders' behavior, Padmanabhan et al. (2021) investigated how individuals changed their bikesharing patterns in three US cities—Boston, New York, and Chicago. The authors found that trip durations for bikeshare users increased during the pandemic. However, bikeshare usage patterns differed in these three cities. Basu and Ferreira (2021) found that car ownership and bikeshare demand increased in Boston during the pandemic, which agrees with the findings from Padmanabhan et al. (2021). All these studies found that demand and trip durations among bikeshare users increased during the pandemic. Also during the pandemic, trip purposes for bikeshare users were more recreational and nonwork related than before the pandemic.

Most pandemic-related studies addressed in this literature review relied on descriptive statistics or elementary modeling techniques (e.g., ordinary least squares). Therefore, using their results to perform advanced forecasting and develop evidence-based policy recommendations is not feasible. This literature review shows a clear methodological gap in research on bikesharing during the pandemic.

*Effect of sociodemographic factors on bikeshare demand*

Several previous studies have examined factors that affect the bikeshare demand in New York City and some other cities in the United States. A few of these studies relied on mathematical models to identify factors contributing to the bikeshare demand, and the others relied on descriptive statistics. Specifically, Reilly et al. (2020a) found that households with lower incomes and fewer vehicles are more likely to choose bikesharing. In another study, they found that the bikesharing patterns of males and females vary depending on how urban or suburban the area is (Reilly et al. 2020b). A set of studies also examined demographic inequalities in bikeshare access and found that a significant number of groups, predominantly minority and low-income, do not have access to bikeshare systems. Babagoli et al. (2019) developed a model using the World Health Organization's Health Economic Assessment Tool (HEAT) and found that the expansion of Citi Bike service directly benefited low-income neighborhoods and communities of color. Furthermore, Ursaki and Aultman-Hall (2015) investigated the unequal distribution of bikeshare programs among different population groups in US cities. Their results show that significant statistical differences exist in bikeshare access depending on income, education level, and race.

*Effect of weather and built environment on bikeshare demand*

A few studies have examined how weather and built environment attributes affect bikeshare demand in various cities. Heaney et al. (2019) found that increasing temperature negatively impacts bikeshare demand. Wang and Noland (2021a) found the same result in their study. El-Assi et al. (2017) found that bikeshare demand decreases with increasing snow, precipitation, and humidity. Generally, any weather condition uncomfortable for a bicyclist tends to negatively affect bikeshare demand. Because many other studies reported the same finding, this paper refrains from mentioning them in this review. A few other studies have examined the effect of the built environment and infrastructure on bikeshare demand. Xu and Chow (2020) found that increasing the number of stations is positively correlated with bikeshare demand. However, this is only true for trips taken on weekdays. De Chardon et al. (2016) found that bikeshare stations near transit hubs experience more demand than other stations and thus more rebalancing.

A few recent studies have used machine learning algorithms to model bikeshare demand. Yang et al. (2018) used convolutional neural networks to predict daily bikeshare demand at the station and city levels and how it is affected by weather variations. They found that in New York City, increasing precipitation gradually decreases bikeshare demand. However, many individuals still bike while precipitation is high. The authors found a more straightforward relationship between snowfall and bikeshare demand: bikeshare demand reduces significantly with increasing snowfall. Albuquerque et al. (2021) performed an in-depth literature review to identify seminal research using machine learning techniques to model bikeshare demand. They also found that weather is a critical factor in bikeshare demand in megacities.

Another notable study developed bike renting and returning models using deep long short-term memory models (Pan et al. 2019). The study used three data types: weather, history of biking, and time. Ashqar et al. (2019) employed the random forest modeling to predict station-level bikeshare demand in the SF Bay Area and found it is affected by temperature, humidity levels, and the time of the day.

A significant number of studies have examined the aggregated demand for bikeshare services in the United States, especially in New York City. However, as mentioned previously in this literature review, many studies analyzing bikeshare demand in New York City have relied on descriptive statistics (Wang and Noland 2021b; Jobe and Griffin 2021; Basu and Ferreira 2021; Babagoli et al. 2019). Findings from descriptive statistics, although they may be important, are specific to the study location. Moreover, many other studies on bikeshare demand have relied on regression models, which tend to ignore correlated variables unless interaction terms are used (Padmanabhan et al. 2021; Reilly et al. 2020a). For all these reasons, advanced mathematical models are needed to forecast and plan well.

**Data sources**

In this study, three data sources were used: CBS data from New York City obtained from the Citi Bike website, station-level historical weather data from the Network for Environment and Weather Applications accessed on July 31, 2021, and official public holiday information listed on the New York City website accessed on July 31, 2022.

The CBS data provide trip-level information (i.e., transaction records) for all Citi Bike trips and are publicly available. Each transaction record includes trip duration; bike check out and check in times; the names, latitudes, and longitudes of the start and end stations; user ID and type (customer or subscriber); and the gender and birth year of the user. CBS data from March 1, 2019 to February 28, 2021—a total of 39.9 million CBS bikesharing transactions—were obtained for this study. This time frame was chosen to allow for two 1 year data strata: pre-pandemic and during the pandemic. (The first COVID-19 case in New York state was identified on March 1, 2020.)

Given the focus of this study, bikesharing transactions were aggregated by the hour to determine citywide demand. The hour of day was defined using a numeric code, where 0 = 12 a.m., 1 = 1 a.m., … 22 = 10 p.m., and 23 = 11 p.m. Numeric codes were also used to define the day of the week (e.g., 1 = Monday) and the month (e.g., 1 = January). After the aggregation, the pre-pandemic data set contained 8,785 records, and the pandemic data set contained 8,637 records. Shares by gender (male and female), age groups (16–24, 25–44, 45–64, and 65+), and user type (customer and subscriber) were calculated as percentages from the CBS data. Trip characteristics such as average great circle distances (GCDs) and trip durations (in minutes) were also estimated for the study data sets. The GCDs were calculated based on the latitudes and longitudes of trip origins and destinations. Weather data used in this study included hourly air temperatures, precipitation, humidity, and wind speeds. Public holidays were coded as a binary indicator variable.

Table 2 presents descriptive statistics on variables included in the pre-pandemic and pandemic data sets. On average, there were approximately 2,394 bikeshare demands (i.e., number of trips) per hour before the pandemic compared with 2,183 per hour during the pandemic. The average shares for female and customer users (i.e., percentages of users who were female and customers, respectively) were slightly higher during the pandemic than before. Also, average trip distance (based on GCDs) and trip duration were higher during the pandemic. The average weather conditions were fairly similar in both data sets.

Figure 1 illustrates the effect of the pandemic on bikesharing by comparing average daily demand in different months. As shown in Figure 1, average daily demand in March 2020 (during the pandemic) was lower than in March 2019 (pre-pandemic). Furthermore, the average daily bikeshare demand for April 2020 was significantly lower than the pre-pandemic demand because of the restrictions on public activity in New York City. However, in October, November, and December, average daily bikeshare demands were higher in 2020 than in 2019.

Figure 2 shows spatial variations in bikeshare demand at the station level before and during the pandemic. Total trips originated from each station before the pandemic are illustrated in Figure 2(a), and total trips terminated at each station before the pandemic are shown in Figure 2(b). Similarly, total trips originated from each station during the pandemic are displayed in Figure 2(c), and total trips terminated at each station are presented in Figure 2(d). Figure 2 shows that the stations in midtown Manhattan had more trips (both originated and terminated), before and during the pandemic, compared to other areas. The comparisons illustrated in Figure 2 show that CBS expanded its service to more customers during the pandemic. Another important pattern revealed in Figure 2 is that stations near Grand Central Terminal had significantly fewer trips, either originated or terminated, during the pandemic than before. This finding appears to correlate with decreased subway ridership in New York City during the pandemic.

**Methodology**

*Machine learning model*

A machine learning modeling framework was used to investigate hourly demand. Specifically, the XGBoost algorithm was used. XGBoost is an improved version of the standard gradient boosting decision tree algorithm that can construct boosted trees efficiently and operate in parallel. For regression problems, it optimizes the values of objective function and implements the algorithm under the framework of gradient boosting. The Python XGBoost library was accessed on July 31, 2022, and used to train and evaluate the models.

*Interpretation of the machine learning model*

The developed machine learning models were interpreted using the SHAP framework. Inspired by cooperative game theory, the SHAP framework treats all variables as contributors. For each predicted data record, the model generates a predicted value, and the SHAP value is the value assigned to each feature (i.e., explanatory variable) in the data. The Python SHAP library was accessed on July 31, 2022, and used to interpret the developed models.

**Results and discussion**

Two XGBoost models were trained and evaluated: a pre-pandemic model developed based on the pre-pandemic data set and a pandemic model developed based on the pandemic data set. For each data set, 80% of the data were randomly selected and used to train the corresponding model, and the remaining 20% were used to test the model. To evaluate the stability of the models' performance, 10-fold cross-validation was performed. In addition, XGBoost hyperparameters were tuned using a grid search procedure. The grid search yielded the following hyperparameters for the pre-pandemic model: maximum depth of a tree, *max_depth* = 7; minimum sum of instance weight needed in a child, *min_child_weight* = 1; step size, *eta* = 0.01; subsample ratio of the training instances, *subsample* = 0.8; and subsample ratio of columns when constructing each tree, *colsample_bytree* = 0.7. The hyperparameters for the pandemic model were *max_depth* = 8, *min_child_weight* = 3, *eta* = 0.01, *subsample* = 0.8, and *colsample_bytree* = 1. The other parameters in both models were the same as the default options included in the XGBoost library. Table 3 presents the results for these models. Based on the $R^2$ and root mean squared log error (RMSLE) values shown at the bottom of Table 3, both models fit the data well.

*Relative importance of explanatory variables*

The contributions of explanatory variables were calculated via XGBoost's *feature importance* score (see Table 3). A higher value of relative importance indicates a stronger contribution to the estimated bikeshare demand. For the pre-pandemic model, the five most important variables are share of female users, hour of day, average GCD, share of users in the age group 65+, and average trip duration. A few other past studies also found distance and gender as important determinants of bikeshare demand (Wang and Noland 2021b; Reilly et al. 2020a). However, unlike the XGBoost model, traditional regression models and discrete choice models do not determine relative importance or rank, although the value of the coefficients (also elasticity) of these models could be treated as a comparable index to relative importance. The contribution rates for the top two variables, share of female users and hour of day, are 24.5% and 21.3%, respectively. For the pandemic model, the five most important variables are share of female users, hour of day, month, average GCD, and precipitation. The contribution rates for the top two variables, share of female users and hour of day, are 47.4% and 19.1%, respectively. Comparing the pre-pandemic and pandemic models, share of female users, hour of day, and average GCD had similar importance, and month had higher importance in the pandemic model than in the pre-pandemic model. Share of subscriber had no importance (0%) in the pandemic model.

Variable importance was also analyzed via SHAP variable importance plots, which list the variables in descending order of importance. Because they contribute more to the model, the variables listed closer to the top of each plot have higher predictive power than those listed closer to the bottom. Figure 3(a) shows the variable importance plot for the pre-pandemic model. The dots in the plot represent the samples used to train the model. The horizontal location of each dot shows whether the effect of the variable value is associated with a higher (right) or lower (left) predictive power. The color shows whether the variable has a high (red) or low (blue) value for that observation. Please refer to the online version of the figure for color legends. For example, Figure 3(a) shows that afternoon hours have a high and positive effect on bikeshare demand.

Figure 3(b) shows how the explanatory variables are correlated with the target variable (i.e., bikeshare demand) before the pandemic. A red bar means the variable is positively correlated with the target variable, whereas a blue bar means it is negatively correlated with the target variable. Hour of day, average GCD, share of female users, average trip duration, temperature, month, share of users in the age group 25–44, and share of subscriber users are positively correlated with bikeshare demand. On the other hand, day of the week, share of users in the age group 65+, relative humidity, share of users in the age group 45–64, share of customer users, share of users in the age group 16–24, share of male users, precipitation, holiday, and wind speed are negatively correlated with bikeshare demand. These results are similar to those of El-Assi et al. (2017), although their study area was Toronto, Ontario, Canada. El-Assi et al. also found that bikeshare demand decreases with increasing precipitation and humidity. In addition, this paper included wind speed as an explanatory variable and found it is negatively correlated with bikeshare demand.

Figure 3(c) and (d) show similar variable importance plots for the pandemic model. These results can be explained similarly to Figure 3(a) and (b). Importantly, average trip duration and bikeshare demand are negatively correlated for the pandemic model, whereas they are positively correlated for the pre-pandemic model.

*Variable association analysis*

SHAP value plots show the linear or nonlinear relationships between each explanatory variable and the target variable more intuitively. Considering the relative importance of explanatory variables, nine variables were analyzed. These nine variables ranked higher in relative importance in both the pre-pandemic and pandemic models. The variables are share of female users, hour of day, average GCD, month, precipitation, share of users in the age group 65+, average trip duration, day of the week, and share of male users. Figure 4 presents SHAP value plots for the pre-pandemic model, and Figure 5 presents SHAP value plots for the pandemic model. Share of female users has a nonlinear relationship with bikeshare demand in both models, as shown in Figures 4(a) and 5(a). However, share of female users interacts with average GCD frequently in the pre-pandemic model, whereas share of female users interacts with hour of day frequently in the pandemic model. The probability of bikeshare demand increases when the share of female users is approximately 22% or higher in the pre-pandemic model and approximately 28% or higher in the pandemic model.

Moreover, hour of day has a nonlinear relationship with bikeshare demand, as shown in Figures 4(b) and 5(b). In both models, daytime hours have positive association with bikeshare demand and nighttime hours have negative association with bikeshare demand. In addition, hour of day interacts with average GCD frequently in both models. Average GCD has an approximately linear relationship with bikeshare demand in the pre-pandemic model but has a nonlinear relationship with bikeshare demand in the pandemic model. The share of users in the age group 65+ has a nonlinear relationship with bikeshare demand in both models, and it interacts with share of female users more frequently. Although average trip duration has a linear relationship with bikeshare demand in the pre-pandemic model, it has a nonlinear relationship with bikeshare demand in the pandemic model. Average trip duration frequently interacts with share of female users in the pre-pandemic model and with month in the pandemic model. Finally, precipitation, day of the week, share of male users, and share of users in the age group 16–24 have nonlinear relationships with bikeshare demand in both models.

*Analysis of individual SHAP value*

The modeling framework applied in this study can provide interpretation for each data sample (i.e., a record in the final data set), as well. Why a sample receives its prediction was examined, along with the contributions of the predictors. Figure 6(a) illustrates the prediction for a randomly selected sample for the pre-pandemic model. The base value (2,385) is the mean prediction. The output value (f(x) = 2,845.46) is the prediction for the sample. Variables that increase the predicted value (i.e., move it to the right) are shown in red, and variables that decrease the predicted value (i.e., move it to the left) are shown in blue. Shares of users in age groups 65+ and 25–44, day of the week, humidity, and hour of day increase the predicted value; share of female users, average GCD, and month decrease the predicted value. Figure 6(b) illustrates the prediction for a randomly selected sample record for the pandemic model. For this sample, month increases the prediction, and hour of day, share of female users, average GCD, and air temperature decrease the prediction.

**Conclusions**

To explore the effect of the COVID-19 pandemic on bikeshare demand, this study used CBS transaction (i.e., trip) records from March 2019 to February 2021 in New York City to develop machine learning–based models for estimating hourly bikeshare demand. Two XGBoost models were developed—one using pre-pandemic data and one using data from during the pandemic—using aggregate bikeshare demand synthesized from 39.9 million trips. This study also implemented a SHAP interpretation framework for these models. Based on relative importance, share of female users and hour of day were the most important variables in both models. However, the month variable had a higher importance in the pandemic model than in the pre-pandemic model. SHAP values were calculated for all explanatory variables. SHAP value plots were analyzed to identify the linear or nonlinear relationship between each explanatory variable and the target variable, and SHAP dependence plots were analyzed to determine the interactions among explanatory variables. Hour of day interacts frequently with average GCD between trip origin and destination in the pre-pandemic model, whereas hour of day interacts frequently with share of female users in the pandemic model.

Females were found to use bikeshare systems more than males during the pandemic than before. It is possibly because females linked bikeshare trips with public transit less often than males before the pandemic (Wang and Akar 2019). This indicates that bikeshare usage among females, in addition to the extension of coverage areas and bikeshare infrastructure, should be carefully considered in the future. On average, bikeshare users traveled farther during the pandemic than before. Using bikesharing instead of private cars or public transit for longer trips could have significant environmental benefits (e.g., lower emissions).

This study has a few limitations. Aggregate demand models generally suffer from a lack of sociodemographic variables. Gender and age variables were incorporated into this study, but income, race, and vehicle ownership information were not available in the data used. Furthermore, this study estimated the pre-pandemic and pandemic models separately. However, the models did not include COVID-19 infection rate, which could have elucidated the relationship between COVID-19 infection and bikeshare demand. In a future study, machine learning–based time series models will be tested to capture the effects of lagged demand and moving average components.

Pan, Y., Zheng, R. C., Zhang, J., and Yao, X. 2019. "Predicting Bike Sharing Demand Using Recurrent Neural Networks." *Procedia Computer Science* 147: 562–566. doi:10.1016/j.procs.2019.01.217

Reilly, K. H., Noyes, P., and Crossa, A. 2020a. "From Non-Cyclists to Frequent Cyclists: Factors Associated with Frequent Bike Share Use in New York City." *Journal of Transport & Health* 16: 100790. doi:10.1016/j.jth.2019.100790

Reilly, K. H., Wang, S. M., and Crossa, A. 2020b. "Gender Disparities in New York City Bike Share Usage." *International Journal of Sustainable Transportation* 16 (3): 237–245. doi:10.1080/15568318.2020.1861393

Ursaki, J., and Aultman-Hall, L. 2015. *Quantifying the Equity of Bikeshare Access in US Cities* (TRC Report 15-011). University of Vermont Transportation Research Center.

Wang, K., and Akar, G. 2019. "Gender Gap Generators for Bike Share Ridership: Evidence from Citi Bike System in New York City." *Journal of Transport Geography* 76: 1–19. doi:10.1016/j.jtrangeo.2019.02.003

Wang, H., and Noland, R. B. 2021a. "Bikeshare and Subway Ridership Changes During the COVID-19 Pandemic in New York City." *Transport Policy*, 106: 262-270. doi:10.1016/j.tranpol.2021.04.004

Wang, H., and Noland, R. 2021b. "Changes in the Pattern of Bikeshare Usage Due to the COVID-19 Pandemic." *Findings.* doi:10.32866/001c.18728

Xu, S. J., and Chow, J. Y. 2020. "A Longitudinal Study of Bike Infrastructure Impact on Bikesharing System Performance in New York City." *International Journal of Sustainable Transportation* 14 (11): 886–902. doi:10.1080/15568318.2019.1645921

Yang, H., Xie, K., Ozbay, K., Ma, Y., and Wang, Z. 2018. "Use of Deep Learning to Predict Daily Usage of Bike Sharing Systems." *Transportation Research Record* 2672 (36), 92–102. doi:10.1177/036119811880135415

Table 1. Summary of relevant literature on bikeshare demand

| Study | Objective | Method | Study area | Findings |
|---|---|---|---|---|
| Wang and Noland (2021a) | To understand the factors affecting bikeshare and subway ridership during the pandemic | Prais-Winsten models | New York City | • Weather and pandemic prevention policies affect both the bikeshare and subway systems.<br>• The pandemic affected subway ridership more than bikeshare ridership. |
| Wang and Noland (2021b) | To analyze bikeshare demand | Descriptive statistics | Borough of Brooklyn in New York City | • Individuals took more recreational and long-distance trips during the pandemic than before it.<br>• The number of work-related trips decreased drastically. |
| Jobe and Griffin (2021) | To understand the bike share systems' reactions to the pandemic | Descriptive statistics | San Antonio, Texas | • Only 5 out of 11 bikeshare companies actively worked to reduce COVID-19 transmission. |
| Padmanabhan et al. (2021) | To investigate how individuals changed their bikeshare patterns | Correlation coefficients and random parameter ordinary least squares model | Boston, New York, and Chicago | • During the pandemic, the trip duration for bikeshare users increased. |
| Basu and Ferreira (2021) | To investigate sustainable mode choices during pandemic | Descriptive statistics | Boston | • Car ownership and bikeshare ridership both increased during the pandemic. |
| Reilly et al. (2020a) | To investigate factors associated with frequent bikeshare users | Logistic regression | New York City | • Male, households with fewer vehicles and lower income are more likely to use bikeshare. |
| Babagoli et al. (2019) | To examine spatial equity and estimate the health effects of Citi Bike | HEAT model to calculate annual mortality effect and annual economic effect | New York City | • After expansion of Citi Bike service, added benefits are observed among low-income neighborhoods and communities of color. |
| Ursaki and Aultman-Hall (2015) | To identify the unequal distribution of bikeshare programs | Census Block–level data in various US cities | United States | • Statistical difference in bikeshare access exists depending on income, education level, and race. |
| Reilly et al. (2020b) | To understand the gender disparities in bikeshare usage | Bivariate logistic regression models and Wilcoxon signed-rank tests | New York City | • Females are more likely to use bikeshare program in low-density neighborhoods. |
| El-Assi et al. (2017) | To understand how weather and built environment attributes affect | Multilevel/Linear Mixed Effects model | Toronto, Canada | • Bikeshare demand decreases with increasing snow, precipitation, and humidity. |

| Study | Objective | Method | Study area | Findings |
|---|---|---|---|---|
| | | | | bikeshare demand |
| Heaney et al. (2019) | To identify the relationship between temperature and bikeshare demand | Nonparametric generalized additive models | New York City | • If the daily mean temperatures keep increasing over the years, it may reduce the popularity of bikeshare services. |
| Xu and Chow (2020) | To identify the effects of bike infrastructure on bikesharing systems | Autoregressive conditional heteroscedasticity model | New York City | • Bikeshare demand increases with an increasing number of bikeshare stations.<br>• The number of stations affects only weekday trips, not weekend trips. |
| De Chardon et al. (2016) | To perform station-level analysis of the bikeshare program and interview bikeshare operators | Rebalanced effective usage and normalized available bikes are calculated for different stations | Multiple cities in the United States | • Stations near transit hubs experience more rebalancing than other stations. |

Table 2. Descriptive statistics

| Variable | Description | Pre-pandemic (n = 8,785) | | | | Pandemic (n = 8,637) | | | |
| --- | --- | --- | --- | --- | --- | --- | --- | --- | --- |
| | | **Mean** | **S.D.** | **Min.** | **Max.** | **Mean** | **S.D.** | **Min.** | **Max.** |
| **Target variable** | | | | | | | | | |
| Demand | Number of total bikeshare trips in a given hour | 2,393.7 | 2,134.7 | 5 | 10,491 | 2,182.5 | 2,132.4 | 1 | 10,201 |
| **Explanatory variables** | | | | | | | | | |
| *User characteristics* | | | | | | | | | |
| Male (%) | Share of male users | 71.7 | 7.0 | 51 | 100 | 60.0 | 13.0 | 3 | 100 |
| Female (%) | Share of female users | 21.2 | 5.2 | 0 | 33 | 24.1 | 7.7 | 0 | 47 |
| Age 16–24 (%) | Share of users in the age group 16 to 24 | 9.4 | 3.9 | 0 | 40 | 8.9 | 5.1 | 0 | 50 |
| Age 25–44 (%) | Share of users in the age group 25 to 44 | 57.9 | 5.5 | 24 | 100 | 53.3 | 12.0 | 0 | 100 |
| Age 45–64 (%) | Share of users in the age group 45 to 64 | 30.3 | 5.8 | 0 | 68 | 35.2 | 13.8 | 0 | 98 |
| Age 65+ (%) | Share of users in the age group 65+ | 2.4 | 1.3 | 0 | 13 | 2.5 | 1.8 | 0 | 33 |
| Customer (%) | Share of 24 h pass or 3 day pass users | 12.9 | 8.4 | 0 | 48 | 26.8 | 16.9 | 0 | 97 |
| Subscriber (%) | Share of annual members | 87.1 | 8.4 | 52 | 100 | 73.2 | 16.9 | 3 | 100 |
| *Trip characteristics* | | | | | | | | | |
| GCD (mi) | Average GCD between trip origins and destinations | 1.0 | 0.1 | 0.4 | 1.4 | 1.2 | 0.1 | 0.4 | 2.3 |
| Trip Duration (minutes) | Average trip durations | 12.8 | 2.3 | 5.3 | 23.5 | 16.2 | 3.6 | 5.9 | 34.7 |
| *Weather and holiday* | | | | | | | | | |
| Temperature (°F) | Air temperature | 56.3 | 16.6 | 15 | 94 | 56.4 | 16.8 | 15 | 95 |
| Precipitation (in.) | Precipitation | 0.01 | 0.04 | 0 | 1.1 | 0.01 | 0.04 | 0 | 1.6 |
| Humidity (%) | Relative humidity | 62.6 | 19.8 | 13 | 100 | 62.4 | 19.4 | 15 | 100 |
| Wind speed (mph) | Wind speed | 5.3 | 4.0 | 0 | 31.1 | 5.9 | 4.1 | 0 | 33.0 |
| Holiday | Public holiday (1 = holiday; 0 = otherwise) | 0.1 | 0.2 | 0 | 1 | 0.1 | 0.2 | 0 | 1 |

Table 3. Model results and comparison

| Variable | Description | Pre-pandemic | | Pandemic | |
|---|---|---|---|---|---|
| | | Relative Importance (%) | Rank | Relative Importance (%) | Rank |
| Hour of day | Hour of day (0: 12 a.m., 1: 1 a.m., …, 23: 11 p.m.) | 21.31 | 2 | 19.05 | 2 |
| Day of week | Day of week (1: Monday, 2: Tuesday, …, 7: Sunday) | 2.76 | 7 | 1.25 | 10 |
| Month | Month (1: January, 2: February, …, 12: December) | 1.24 | 15 | 9.96 | 3 |
| Male (%) | Share of male users | 2.58 | 8 | 1.39 | 8 |
| Female (%) | Share of female users | 24.52 | 1 | 47.37 | 1 |
| Age 16–24 (%) | Share of users in the age group 16 to 24 | 2.24 | 9 | 1.65 | 7 |
| Age 25–44 (%) | Share of users in the age group 25 to 44 | 1.27 | 13 | 0.67 | 15 |
| Age 45–64 (%) | Share of users in the age group 45 to 64 | 1.27 | 14 | 0.74 | 12 |
| Age 65+ (%) | Share of users in the age group 65+ | 10.79 | 4 | 3.49 | 6 |
| Customer (%) | Share of 24 h pass or 3 day pass users | 1.31 | 12 | 1.30 | 9 |
| Subscriber (%) | Share of annual members | 1.55 | 11 | 0.00 | 18 |
| GCD (mi) | Average of great circle distances between trip origins and destinations | 19.49 | 3 | 6.39 | 4 |
| Trip duration (minutes) | Average trip durations | 3.03 | 5 | 0.68 | 14 |
| Temperature (°F) | Air temperature | 0.98 | 16 | 1.18 | 11 |
| Precipitation (in.) | Precipitation | 2.93 | 6 | 3.53 | 5 |
| Humidity (%) | Relative humidity | 0.71 | 17 | 0.68 | 17 |
| Wind speed (mph) | Wind speed | 0.16 | 18 | 0.17 | 13 |
| Holiday | Public holiday (1 = holiday; 0 = otherwise) | 1.86 | 10 | 0.50 | 16 |
| $R^2$ | | | 0.985 | | 0.971 |
| RMSLE | | | 0.312 | | 0.368 |

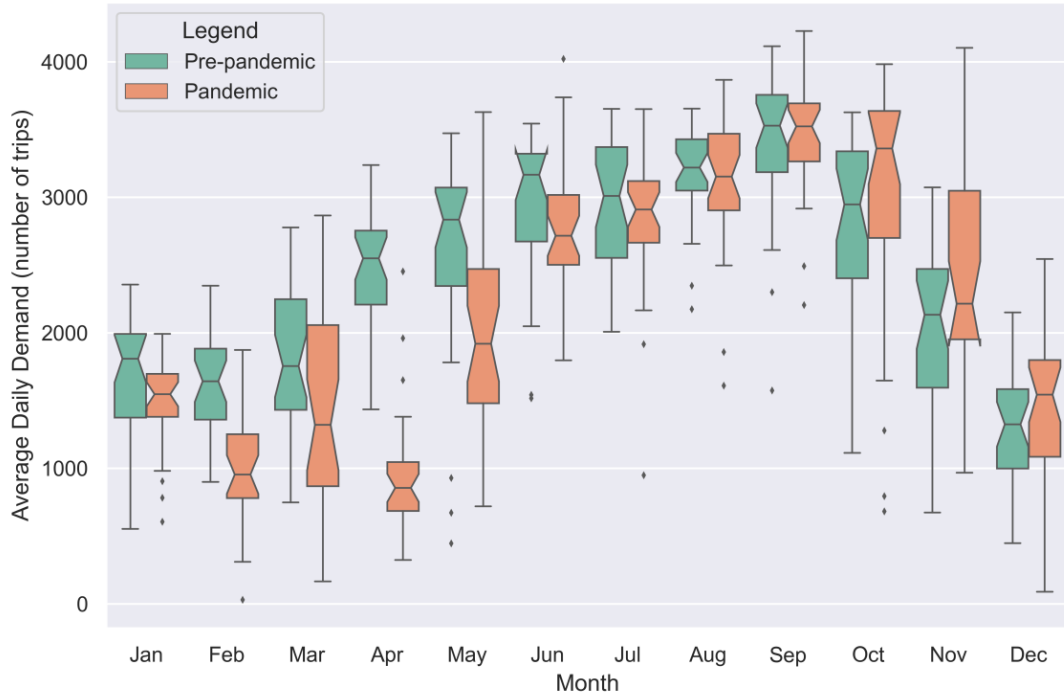

Figure 1. Monthly variation in average daily bikeshare demand before and during the pandemic.

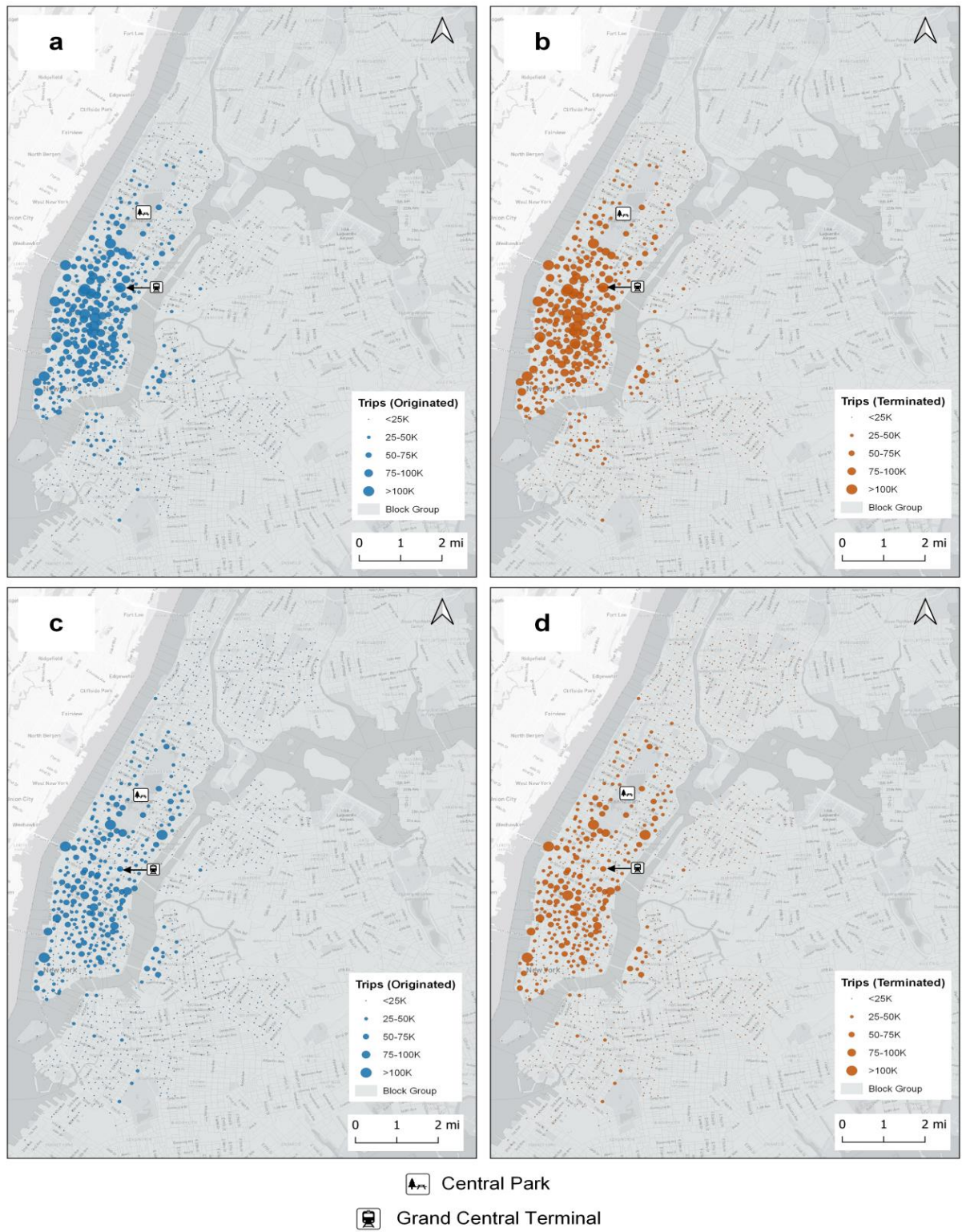

Figure 2. Spatial variations in bikeshare demand at the station level in New York City: (a) trips originated before the pandemic, (b) trips terminated before the pandemic, (c) trips originated during the pandemic, (d) trips terminated during the pandemic.

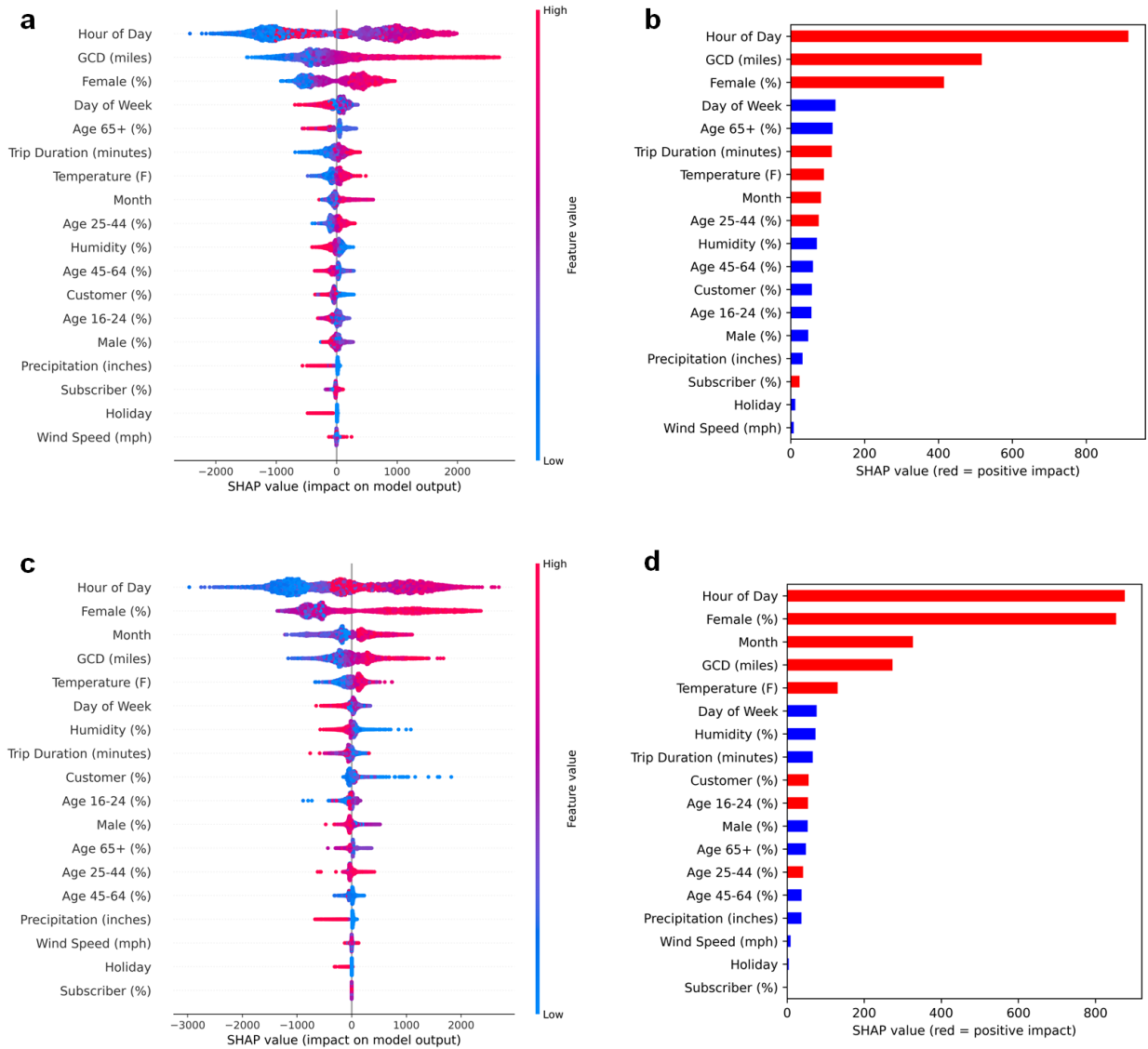

Figure 3. Model interpretation: (a) SHAP variable importance plot for the pre-pandemic model and (b) a simplified version of the plot; (c) SHAP variable importance plot for the pandemic model and (d) a simplified version of the plot.

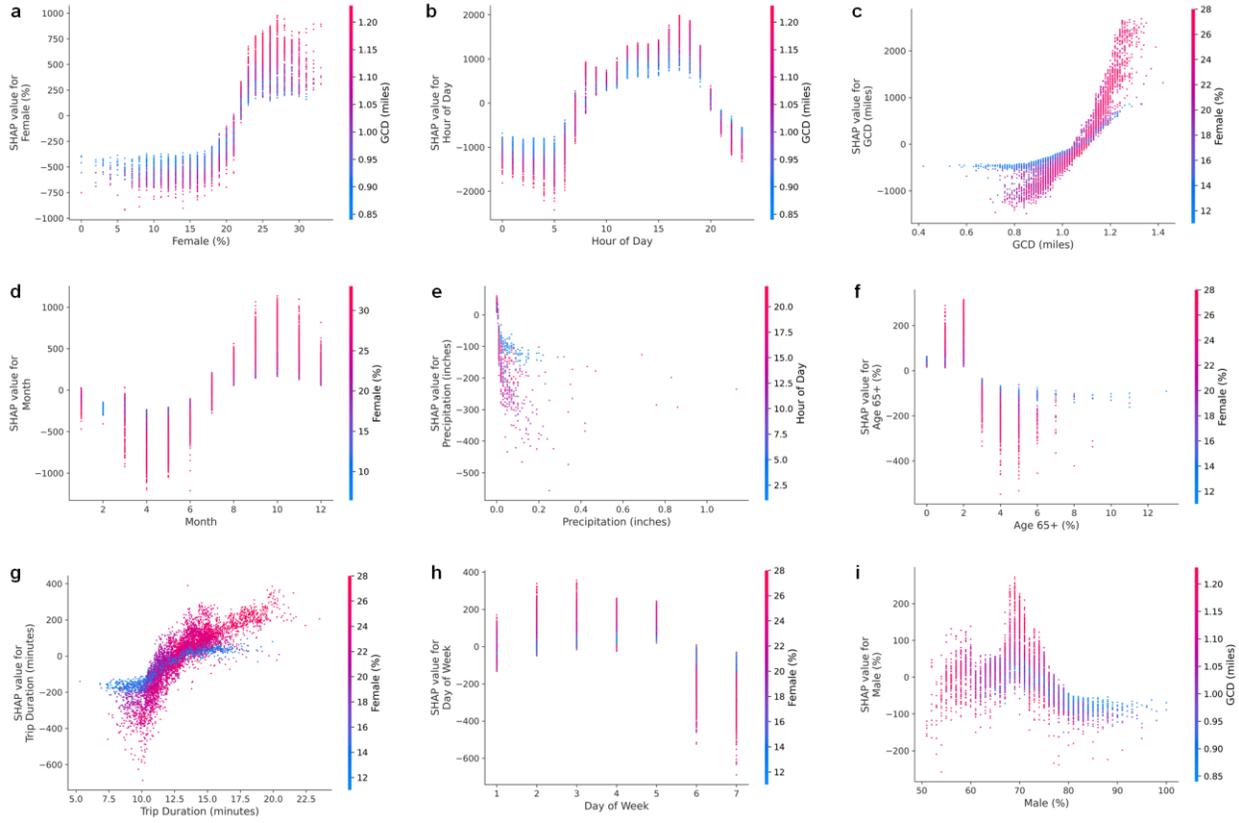

Figure 4. SHAP value plot of top-ranked variables for pre-pandemic model.

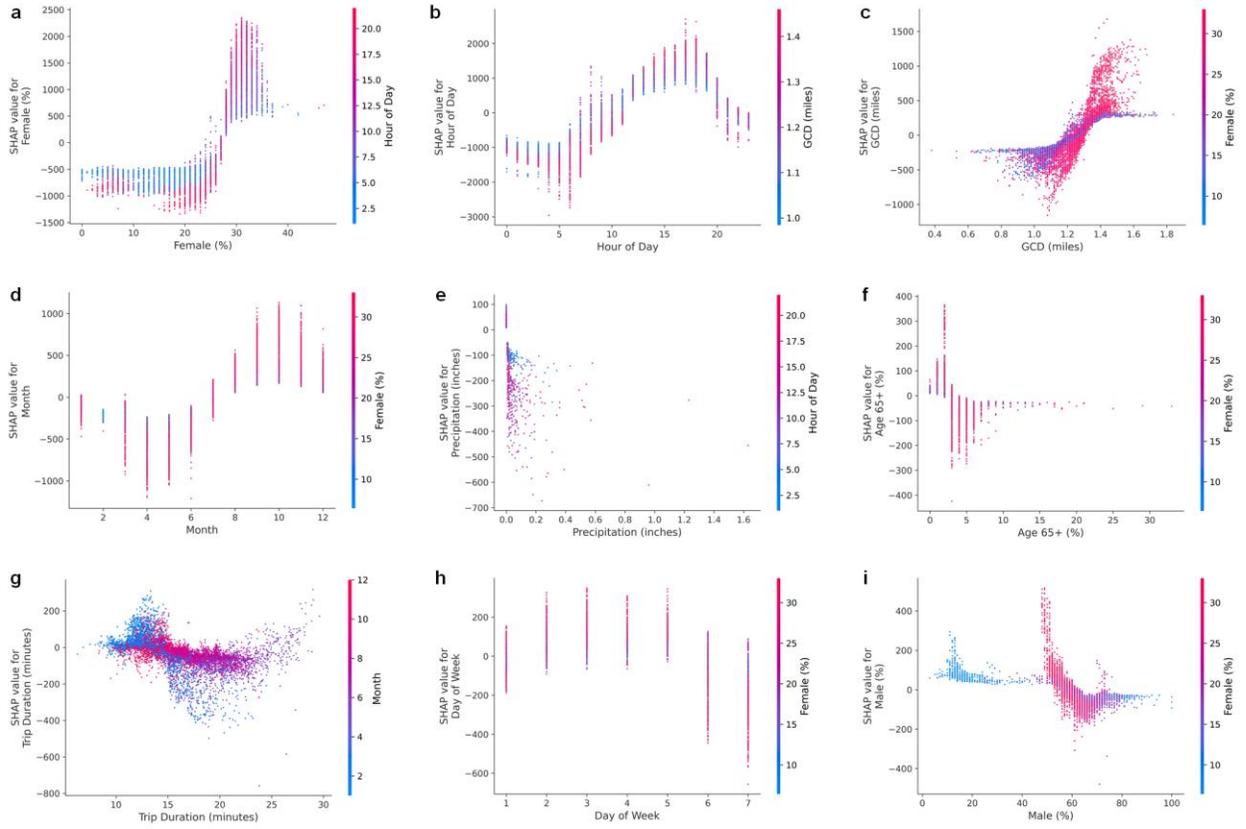

Figure 5. SHAP value plot of top-ranked variables for pandemic model.

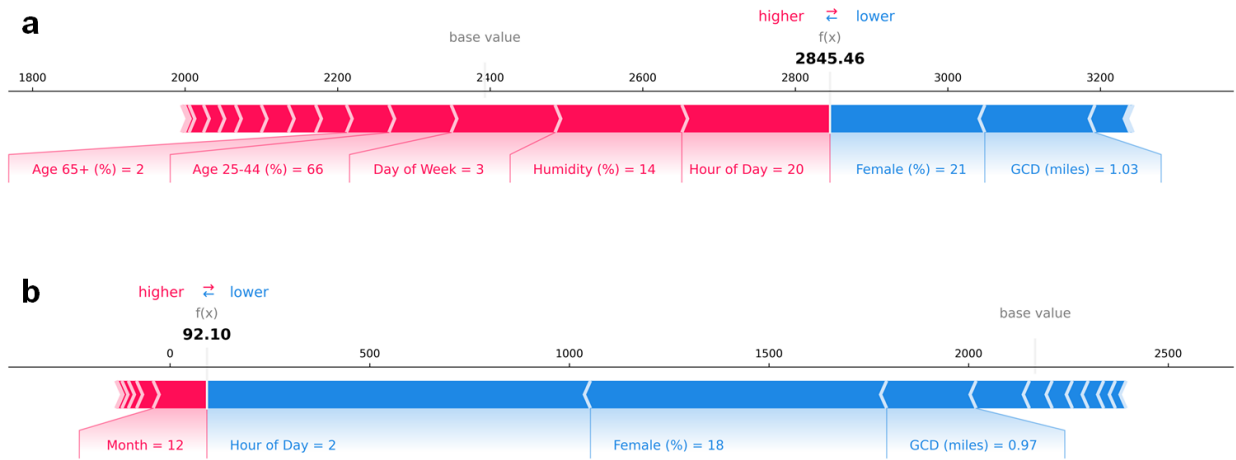

Figure 6. Individual SHAP value plots based on randomly selected samples for the (a) pre-pandemic model and (b) pandemic model.